\documentclass[11pt,a4paper]{article}
\usepackage[breaklinks=true]{hyperref}
\usepackage[hyperref]{acl2017}
\usepackage{times}
\usepackage{latexsym}
\usepackage{verbatim}
\usepackage{xspace}
\usepackage{textcomp}
\usepackage[small]{caption}
\usepackage{graphics}
\usepackage{tabularx}

\usepackage{url}
\usepackage{xcolor}

\aclfinalcopy 
\def\aclpaperid{3584} 

\newcommand{\rnngd}{\textsc{RD}\xspace}
\newcommand{\lstm}{\textsc{LM}\xspace}
\newcommand{\rnngg}{\textsc{RG}\xspace}
\newcommand{\p}{\ensuremath{\rightarrow}\xspace}
\newcommand{\un}{\ensuremath{\cup}\xspace}

\title{Improving Neural Parsing by \\ Disentangling Model Combination and Reranking Effects}

\author{
Daniel Fried\thanks{\enskip Equal contribution.} \qquad Mitchell Stern\footnotemark[1] \qquad Dan Klein \\
Computer Science Division \\
University of California, Berkeley \\
{\tt \{dfried,mitchell,klein\}@cs.berkeley.edu}
}

\date{}

\begin{document}

\maketitle

\begin{abstract}

Recent work has proposed several generative neural models for constituency parsing that achieve state-of-the-art results. Since direct search in these generative models is difficult, they have primarily been used to rescore candidate outputs from base parsers in which decoding is more straightforward.  We first present an algorithm for direct search in these generative models.  We then demonstrate that the rescoring results are at least partly due to implicit model combination rather than reranking effects.  Finally, we show that explicit model combination can improve performance even further, resulting in new state-of-the-art numbers on the PTB of 94.25 F1 when training only on gold data and 94.66 F1 when using external data.

\end{abstract}

\section{Introduction}
\label{sec:introduction}

Recent work on neural constituency parsing \citep{dyer2016recurrent, Choe16Parsing}
has found multiple cases where generative \emph{scoring models} for which inference is complex outperform \emph{base models} for which inference is simpler. Let A be a parser that we want to parse with (here one of the generative models), and let B be a base parser that we use to propose candidate parses which are then scored by the less-tractable parser A. We denote this cross-scoring setup by \mbox{B \p A}.
The papers above repeatedly saw that the cross-scoring setup \mbox{B \p A} under which their generative models were applied outperformed the standard single-parser setup \mbox{B \p B}. We term this a \emph{cross-scoring gain}.

This paper asks two questions.  First, \emph{why} do recent discriminative-to-generative cross-scoring setups \mbox{B \p A} outperform their base parsers B?  Perhaps generative models A are simply superior to the base models B and direct generative parsing (\mbox{A \p A}) would be better still if it were feasible. If so, we would characterize the cross-scoring gain from \mbox{B \p B} to \mbox{B \p A} as a \emph{reranking gain}. However, it's also possible that the hybrid system \mbox{B \p A} shows gains merely from subtle model combination effects. If so, scoring candidates using some combined score \mbox{A + B} would be even better, which we would characterize as a \emph{model combination gain}.  It might even be the case that B is a better parser overall (i.e.\ \mbox{B \p B} outperforms \mbox{A \p A}).

Of course, many real hybrids will exhibit both reranking and model combination gains. In this paper, we present experiments to isolate the degree to which each gain occurs for each of two state-of-the-art generative neural parsing models: the Recurrent Neural Network Grammar generative parser (RG) of \citet{dyer2016recurrent}, and the LSTM language modeling generative parser (LM) of \citet{Choe16Parsing}.

In particular, we present and use a beam-based search procedure with an augmented state space that can search directly in the generative models, allowing us to explore \mbox{A \p A} for these generative parsers A independent of any base parsers. Our findings suggest the presence of model combination effects in both generative parsers: when parses found by searching directly in the generative parser are added to a list of candidates from a strong base parser (the RNNG discriminative parser, RD \citep{dyer2016recurrent}), performance decreases when compared to using just candidates from the base parser, i.e.,\ \mbox{B \un A \p A} has lower evaluation performance than \mbox{B \p A} (Section~\ref{sec:augmenting}).

This result suggests that both generative models benefit from fortuitous search errors in the rescoring setting -- there are trees with higher probability under the generative model than any tree proposed by the base parser, but which would decrease evaluation performance if selected. Because of this, we hypothesize that model combination effects between the base and generative models are partially responsible for the high performance of the generative reranking systems, rather than the generative model being generally superior.

Here we consider our second question: if cross-scoring gains are at least partly due to implicit model combination, can we gain even more by combining the models explicitly?  We find that this is indeed the case: simply taking a weighted average of the scores of both models when selecting a parse from the base parser's candidate list improves over using only the score of the generative model, in many cases substantially (Section~\ref{sec:score_combination}). Using this technique, in combination with ensembling, we obtain new state-of-the-art results on the Penn Treebank: 94.25 F1 when training only on gold parse trees and 94.66 F1 when using external silver data.

\section{Decoding in generative neural models}

All of the parsers we investigate in this work (the discriminative parser RD, and the two generative parsers RG and LM, see Section~\ref{sec:introduction}) produce parse trees in a depth-first, left-to-right traversal, using the same basic \emph{actions}: 
\textsc{NT($X$)}, which opens a new constituent with the non-terminal symbol $X$;
\mbox{\textsc{Shift / Gen($w$)}}, which adds a word;
and \textsc{Reduce}, which closes the current constituent. 
We refer to \citet{dyer2016recurrent} for a complete description of these actions, and the constraints on them necessary to ensure valid parse trees.\footnote{The action space for LM differs from RG in two ways: 1) LM has separate reduce actions \textsc{Reduce($X$)} for each non-terminal $X$, and 2) LM allows any action to have non-zero probability at all times, even those that may be structurally invalid. 
}

The primary difference between the actions in the discriminative and generative models is that, whereas the discriminative model uses a \textsc{Shift} action which is fixed to produce the next word in the sentence, the generative models use \textsc{Gen($w$)} to define a distribution over all possible words $w$ in the lexicon. This stems from the generative model's definition of a joint probability $p(x, y)$ over all possible sentences $x$ and parses $y$. To use a generative model as a parser, we are interested in finding the maximum probability parse for a given sentence. This is made more complicated by not having an explicit representation for $p(y | x)$, as we do in the discriminative setting. However, we can start by applying similar approximate search procedures as are used for the discriminative parser, constraining the set of actions such that it is only possible to produce the observed sentence: i.e.\ only allow a \textsc{Gen($w$)} action when $w$ is the next terminal in the sentence, and prohibit \textsc{Gen} actions if all terminals have been produced.

\subsection{Action-synchronous beam search}
\label{sec:action-level-beam}
Past work on discriminative neural constituency parsers has shown the effectiveness of beam search with a small beam \cite{vinyals2015grammar} or even greedy search, as in the case of RD \cite{dyer2016recurrent}. The standard beam search procedure, which we refer to as \emph{action-synchronous}, maintains a beam of $K$ partially-completed parses that all have the same number of actions taken. At each stage, a pool of successors is constructed by extending each candidate in the beam with each of its possible next actions. The $K$ highest-probability successors are chosen as the next beam.

Unfortunately, we find that action-synchronous beam search breaks down for both generative models we explore in this work, failing to find parses that are high scoring under the model. This stems from the probabilities of the actions \textsc{NT($X$)} for all labels $X$ almost always being greater than the probability of \textsc{Gen($w$)} for the particular word $w$ which must be produced next in a given sentence. Qualitatively, the search procedure prefers to open constituents repeatedly up until the maximum number allowed by the model. 
While these long chains of non-terminals will usually have lower probability than the correct sequence at the point where they finally generate the next word, they often have higher probability up until the word is generated, and so they tend to push the correct sequence off the beam before this point is reached.
This search failure produces very low evaluation performance: with a beam of size $K = 100$, action-synchronous beam search achieves 29.1 F1 for RG and 27.4 F1 for LM on the development set.

\subsection{Word-synchronous beam search}
\label{sec:word-level-beam}
To deal with this issue, we force partial parse candidates to compete with each other on a word-by-word level, rather than solely on the level of individual actions. The \emph{word-synchronous} beam search we apply 
is very similar to approximate decoding procedures developed for other generative models \cite{henderson2003inducing,titov2010latent,buys2015generative} and can be viewed as a simplified version of the procedure used in the generative top-down parsers of \citet{roark2001probabilistic} and \citet{charniak2010top}.

In word-synchronous search, we augment the beam state space, identifying beams by tuples $(|W|, |A_w|)$, where $|W|$ is the number of words that have been produced so far in the sentence, and $|A_w|$ is the number of structural actions that have been taken since the last word was produced. Intuitively, we want candidates with the same $|W|=w$ to compete against each other. For a beam of partial parses in the state $(|W|=w, |A_w|=a)$, we generate a beam of successors by taking all of the next possible actions for each partial parse in the beam. If the action is \textsc{NT}($X$) or $\textsc{Reduce}$, we place the resulting partial parse in the beam for state $(|W|=w, |A_w|=a+1)$; otherwise, if the action is  \textsc{Gen}, we place it in a list for $(|W|=w+1, |A_w|=0)$. After all partial parses in the beam have been processed, we check to see if there are a sufficient number of partial parses that have produced the next word: if the beam $(|W|=w+1, |A_w|=0)$ contains at least $K_w$ partial parses (the \emph{word beam size}), we prune it to this size and continue search using this beam. Otherwise, we continue building candidates for this word by pruning the beam $(|W|=w, |A_w|=a+1)$ to size $K_a$ (the \emph{action beam size}), and continuing search from there.

\begin{table}
\centering
\begin{tabular}{r|ccccccc}
& \multicolumn{6}{c}{Word-synchronous beam size, $K_w$} \\
model &  10 & 20 & 40 & 60 & 80 & 100 \\
\hline
\rnngg & 74.1 & 80.1 & 85.3 & 87.5 & 88.7 & 89.6\\
\lstm & 83.7 & 88.6 & 90.9 & 91.6 & 92.0 & 92.2 \\
\end{tabular}
\caption{\label{tab:beam_size}F1 on the development set for word-synchronous beam search when searching in the RNNG generative (RG) and LSTM generative (LM) models. $K_a$ is set to $10 \times K_w$.}
\vspace{-1em}
\end{table}

In practice, we found it to be most effective to use a value for $K_w$ that is a fraction of the value for $K_a$. In all the experiments we present here, we fix $K_a = 10 \times K_w$, with $K_w$ ranging from 10 to 100.
Table~\ref{tab:beam_size} shows F1 for decoding in both generative models on the development set, using the top-scoring parse found for a sentence when searching with the given beam size. RG has comparatively larger gains in performance between the larger beam sizes, while still underperforming LM, suggesting that more search is necessary in this model.

\section{Experiments}

Using the above decoding procedures, we attempt to separate reranking effects from model combination effects through a set of reranking experiments. Our base experiments are performed on the Penn Treebank~\cite{marcus1993building}, using sections \mbox{2-21} for training, section 22 for development, and section 23 for testing. For the LSTM generative model (LM), we use the pre-trained model released by \citet{Choe16Parsing}. We train RNNG discriminative (RD) and generative (RG) models, following \citet{dyer2016recurrent} by using the same hyperparameter settings, and using pre-trained word embeddings from \citet{ling2015two} for the discriminative model. The automatically-predicted part-of-speech tags we use as input for RD are the same as those used by \citet{Cross16Span}.

In each experiment, we obtain a set of candidate parses for each sentence by performing beam search in one or more parsers. We use  action-synchronous beam search (Section \ref{sec:action-level-beam}) with beam size $K = 100$ for RD and word-synchronous beam (Section \ref{sec:word-level-beam}) with $K_w = 100$ and $K_a = 1000$ for the generative models RG and LM. 

In the case that we are using only the scores from a single generative model to rescore candidates taken from the discriminative parser, this setup is close to the reranking procedures originally proposed for these generative models. For RG, the original work also used RD to produce candidates, but drew samples from it, whereas we use a beam search to approximate its $k$-best list. 
The LM generative model was originally used to rerank a 50-best list taken from the Charniak parser~\citep{Charniak:2000:MP:974305.974323}. In comparison, we found higher performance for the LM model when using a candidate list from the RD parser: 93.66 F1 versus 92.79 F1 on the development data. This may be attributable to having a stronger set of candidates: with beam size 100, RD has an oracle F1 of 98.2, compared to 95.9 for the 50-best list from the Charniak parser.

\begin{table}
\centering
\begin{tabular}{r|ccc}
& \multicolumn{3}{c}{Scoring models} \\
Candidates & \rnngd & \rnngg & \rnngd+ \rnngg \\
\hline
\rnngd & 92.22 & 93.45 & 93.87 \\
\rnngg & 90.24 & 89.55 & 90.53 \\
\rnngd \un \rnngg & 92.22 & 92.78 & 93.92  \\
\end{tabular}

\vspace{5pt}

\begin{tabular}{r|ccc}
& \multicolumn{3}{c}{Scoring models} \\
Candidates & \rnngd & \lstm & \rnngd+ \lstm \\
\hline
\rnngd & 92.22 & 93.66 & 93.99 \\
\lstm & 92.57 & 92.20 & 93.07 \\
\rnngd \un \lstm & 92.24 & 93.47 & 94.15  \\
\end{tabular}
\caption{\label{tab:lstm_model_score_combos} Development F1 scores on section 22 of the PTB when using various models to produce candidates and to score them. $\cup$ denotes taking the union of candidates from each of two models; $+$ denotes using a weighted average of the models' log-probabilities.
\vspace{-1em}
}
\end{table}

\subsection{Augmenting the candidate set}
\label{sec:augmenting}
We first experiment with combining the candidate lists from multiple models,  which allows us to look for potential model errors and model combination effects. Consider the standard reranking setup \mbox{B \p A}, where we search in B to get a set of candidate parses for each sentence, and choose the top scoring candidate from these under A. We extend this by also searching directly in A to find high-scoring candidates for each sentence, and combining them with the candidate list proposed by B by taking the union, \mbox{A \un B}. We then choose the highest scoring candidate from this list under A.
If A generally prefers parses outside of the candidate list from B, but these decrease evaluation performance (i.e., if \mbox{B \un A \p A} is worse than \mbox{B \p A}), this suggests a model combination effect is occurring: A makes errors which are hidden by having a limited candidate list from B.

This does seem to be the case for both generative models, as shown in Table~\ref{tab:lstm_model_score_combos}, which presents F1 scores on the development set when varying the models used to produce the candidates and to score them. Each row is a different candidate set, where the third row in each table presents results for the augmented candidate sets; each column is a different scoring model,  where the third column is the \emph{score combination} setting described below. Going from \mbox{RD \p RG} to the augmented candidate setting \mbox{RD \un RG \p RG} decreases performance from 93.45 F1 to 92.78 F1 on the development set. This difference is statistically significant at the $p < 0.05$ level under a paired bootstrap test. We see a smaller, but still significant, effect in the case of LM: \mbox{RD \p LM} achieves 93.66, compared to 93.47 for \mbox{RD \un LM \p LM}.

We can also consider the performance of \mbox{RG \p RG}  and \mbox{LM \p LM} (where we do not use candidates from RD at all, but return the highest-scoring parse from searching directly in one of the generative models) as an indicator of reranking effects: absolute performance is higher for LM (92.20 F1) than for RG (89.55).
Taken together, these results suggest that model combination contributes to the success of both models, but to a larger extent for RG. A reranking effect may be a larger contributor to the success of LM, as this model achieves stronger performance on its own for the described search setting.

\subsection{Score combination} 
\label{sec:score_combination}
If the cross-scoring setup exhibits an implicit model combination effect, where strong performance results from searching in one model and scoring with the other, we might expect substantial further improvements in performance by explicitly combining the scores of both models. To do so, we score each parse by taking a weighted sum of the log-probabilities assigned by both models \cite{hayashi2013efficient}, using an interpolation parameter which we tune to maximize F1 on the development set. 

These results are given in columns \mbox{RD + RG} and \mbox{RD + LM} in Table~\ref{tab:lstm_model_score_combos}. 
We find that combining the scores of both models improves on using the score of either model alone, regardless of the source of candidates. These improvements are statistically significant in all cases. Score combination also more than compensates for the decrease in performance we saw previously when adding in candidates from the generative model: \mbox{RD \un RG \p RD + RG} improves upon both \mbox{RD \p RG} and \mbox{RD \un RG \p RG}, and the same effect holds for LM.

\subsection{Strengthening model combination}
\begin{table}
\centering
\begin{tabularx}{1.05\linewidth}{Xcc}
Model & PTB & +S \\
\hline \hline
\mbox{}\citet{Liu16ShiftReduce} & 91.7 & -- \\
\mbox{}\citet{dyer2016recurrent}-discriminative & 91.7 & -- \\
\mbox{}\citet{dyer2016recurrent}-generative & 93.3 & -- \\
\mbox{}\citet{Choe16Parsing} & 92.6 & 93.8 \\
\hline 
Model Combination \\
\hline
1) \small{\rnngd \p \rnngd}  & 91.51 & 91.73  \\
2) \small{\rnngd \p \rnngg}  & 92.73 & 93.29 \\
3) \small{\rnngd \p \rnngd + \rnngg} & 93.27 & 93.64 \\
4) \small{\rnngd \un \rnngg \p\rnngd + \rnngg} &93.45 & 93.75  \\
5) \small{\rnngd \p \lstm}  & 93.31 & 94.18  \\
6) \small{\rnngd \p \rnngd + \lstm} &93.71 & 94.27 \\
7) \small{\rnngd \un \lstm \p \rnngd + \lstm}& 93.89 & 94.63 \\
8) \small{\rnngd \p \rnngd + \rnngg + \lstm}&93.63 & 94.33 \\
9) \small{\rnngd \un \rnngg \un \lstm \p \rnngd + \rnngg + \lstm} &{\bf 93.94} & {\bf 94.66}\\
\hline
Ensembling \\
\hline
10) \small{\rnngd(8) \p \rnngd(8)} & 92.72 & 92.53 \\
11) \small{\rnngd(8) \p \rnngd(8) + \rnngg(8)}  & 94.09 & 94.22 \\
12) \small{\rnngd(8) \p \rnngd(8) + \lstm}  & 93.97 & 94.56  \\
13) \small{\rnngd(8) \p \rnngd(8) + \rnngg(8) + \lstm} & {\bf 94.25} & {\bf 94.62} \\

\end{tabularx}
\caption{\label{tab:final_test_results} Test F1 scores on section 23 of the PTB, by treebank training data conditions: either using only the training sections of the PTB, or using additional silver data (+S).}
\vspace{-1em}
\end{table}

Given the success of model combination between the base model and a single generative model, we also investigate the hypothesis that the generative models are complementary. The Model Combination block of Table~\ref{tab:final_test_results} shows full results on the test set for these experiments, in the PTB column. The same trends we observed on the development data, on which the interpolation parameters were tuned, hold here: score combination improves results for all models (row 3 vs.\ row 2; row 6 vs.\ row 5), with candidate augmentation from the generative models giving a further increase (rows 4 and 7).\footnote{These increases, from adding score combination and candidate augmentation, are all significant with $p < 0.05$ in the PTB setting. In the +S data setting, all are significant except for the difference between row 5 and row 6.} Combining candidates and scores from all three models (row 9), we obtain 93.94 F1.

\paragraph{Semi-supervised silver data}
\mbox{}\citet{Choe16Parsing} found a substantial increase in performance by training on external data in addition to trees from the Penn Treebank. This \emph{silver dataset} was obtained by parsing the entire New York Times section of the fifth Gigaword corpus using a product of eight Berkeley parsers~\citep{Petrov10Products} and ZPar~\citep{Zhu13Fast}, then retaining 24~million sentences on which both parsers agreed.
For our experiments we train RD and RG using the same silver dataset.\footnote{When training with silver data, we use a 1-to-1 ratio of silver data updates per gold data updates, which we found to give significantly faster convergence times on development set perplexity for RD and RG compared to the 10-to-1 ratio used by \citet{Choe16Parsing} for LM. 
}
The +S column in Table~\ref{tab:final_test_results} shows these results, where we observe gains over the PTB models in nearly every case. As in the PTB training data setting, using all models for candidates and score combinations is best, achieving 94.66 F1 (row 9).

\paragraph{Ensembling}
Finally, we compare to another commonly used model combination method: ensembling multiple instances of the same model type trained from different random initializations. We train ensembles of 8 copies each of RD and RG in both the PTB and silver data settings, combining scores from models within an ensemble by averaging the models' distributions for each action (in beam search as well as rescoring). These results are shown in the bottom section, Ensembling, of Table~\ref{tab:final_test_results}. 

Performance when using only the ensembled RD models (row 10) is lower than rescoring a single RD model with score combinations of single models, either \mbox{RD + RG} (row 3) or \mbox{RD + LM} (row 6). In the PTB setting, ensembling with score combination achieves the best overall result of 94.25 (row 13).  In the silver training data setting, while this does improve on the analogous unensembled result (row 8), it is not better than the combination of single models when candidates from the generative models are also included (row~9).

\section{Discussion}
Searching directly in the generative models yields results that are partly surprising, as it reveals the presence of parses which the generative models prefer, but which lead to lower performance than the candidates proposed by the base model. However, the results are also unsurprising in the sense that explicitly combining scores allows the reranking setup to achieve better performance than implicit combination, which uses only the scores of a single model. Additionally, we see support for the hypothesis that the generative models can achieve good results on their own, with the LSTM generative model showing particularly strong and self-contained performance.  

While this search procedure allows us to explore these generative models, disentangling reranking and model combination effects, the increase in performance from augmenting the candidate lists with the results of the search may not be worth the required computational cost in a practical parser. However, we do obtain a gain over state-of-the-art results using simple model score combination on only the base candidates, which can be implemented with minimal cost over the basic reranking setup. This provides a concrete improvement for these particular generative reranking procedures for parsing. More generally, it supports the idea that hybrid systems, which rely on one model to produce a set of candidates and another to determine which candidates are good, should explore combining their scores and candidates when possible.

\section*{Acknowledgments}

We would like to thank Adhiguna Kuncoro and Do Kook Choe for their help providing data and answering questions about their work, as well as Jacob Andreas, John DeNero, and the anonymous reviewers for their suggestions. DF is supported by an NDSEG fellowship. MS is supported by an NSF Graduate Research Fellowship.

\bibliography{refs}
\bibliographystyle{acl_natbib}

\end{document}